# A task of anomaly detection for a smart satellite Internet of things system

Zilong Shao [1, a], Xiaoyu Qu [2, b*], and Fanyang Bu [3, c]

[1]School of Mathematics, Hefei University of Technology, Anhui, China.

[2] School of Electrical and Automation Engineering, Hefei University of Technology, Anhui, China.

[3]CCNU Wollongong Joint Institute, Central China Normal University, Hubei, China.

[a] 2022180181@mail.hfut.edu.cn, [b] 2020214054@mail.hfut.edu.cn, [c] c576395764@126.com

**Abstract.** When the equipment is working, real-time collection of environmental sensor data for anomaly detection is one of the key links to prevent industrial process accidents and network attacks and ensure system security. However, under the environment with specific real-time requirements, the anomaly detection for environmental sensors still faces the following difficulties: (1) The complex nonlinear correlation characteristics between environmental sensor data variables lack effective expression methods, and the distribution between the data is difficult to be captured. (2) it is difficult to ensure the real-time monitoring requirements by using complex machine learning models, and the equipment cost is too high. (3) Too little sample data leads to less labeled data in supervised learning. This paper proposes an unsupervised deep learning anomaly detection system. Based on the generative adversarial network and self-attention mechanism, considering the different feature information contained in the local subsequences, it automatically learns the complex linear and nonlinear dependencies between environmental sensor variables, and uses the anomaly score calculation method combining reconstruction error and discrimination error. It can monitor the abnormal points of real sensor data with high real-time performance and can run on the intelligent satellite Internet of things system, which is suitable for the real working environment. Anomaly detection outperforms baseline methods in most cases and has good interpretability, which can be used to prevent industrial accidents and cyber-attacks for monitoring environmental sensors.

## 1. Introduction

Internet of Things (IoT) as the basic carrier of contemporary information society, the Internet connects people, and the requirements for communication in the information society are not limited to people. The development of Internet of Things (IoT) technology extends communication to people and things, things and things. IoT devices refer to devices that can communicate with each other and do not require direct human involvement [1]. Such as smart home devices, smart wearable devices, temperature/humidity sensors, etc. An anomaly is usually defined as points in certain time steps where the system's behaviors is significantly different from the previous normal status [2]. Anomaly detection is one of the key links to prevent industrial process accidents and ensure system safety. Safe and reliable anomaly detection mechanism can greatly reduce the risk of system failure or unexpected shutdown [3].

In recent years, environmental monitoring and anomaly detection have become increasingly important in industry, agriculture, urban planning, and health. With the development of the Internet of Things, the wide application of environmental sensors provides us with real-time monitoring and data collection capabilities for environmental conditions and parameters. The combination between satellites and IoT presents many exciting developments. This combination has led to new opportunities and innovative solutions in many areas. Satellites provide a wide coverage area for IoT. Traditional IoT relies on terrestrial infrastructure, and coverage is usually limited by the geographical lo-cation of sensors or devices. By combining with satellite networks, IoT can achieve global connectivity, whether in remote areas, oceans, aviation, or other places that are difficult to cover, enabling real-time data transmission and remote monitoring. The combination of satellites and IoT offers new opportunities for connectivity and data exchange on a global scale. It promotes the development of the Internet of things and brings a broader space for innovation in many industries and fields.

Traditional rule-based or threshold-based methods may work in some cases but are often very limiting for complex environmental patterns and changes. Literature [4, 5, 6, 7, 8, 9] performs device identification based on traditional supervised learning, and their adopted models often include AdaBoost, KNN, decision tree, random forest, naive Bayes, and SVM, among others. In this case, with the increasing complexity and dimensionality of sensor data, it may be too much for humans to manually monitor these data and make classification and judgment. Therefore, there is a need for better anomaly detection methods that can quickly detect anomalies in multivariate data with high real-time performance.

## 2. Anomaly detection in smart satellite IoT systems

### 2.1 Architecture of smart satellite Internet of Things

Satellite Internet of things technology refers to the use of satellite broadband communication to provide Internet of things services and realize the integration of space-ground and people-things connection and information interaction. By analyzing and processing information, the management, control, and decision-making of the wider physical world are realized. Its technical basis is satellite communication technology, which realizes communication between two or more terminals by forwarding radio waves through artificial earth satellites as relay stations. The core advantage of satellite Internet of things lies in its global coverage and stability, which can achieve wide coverage, low delay, broadband, low cost and other characteristics.

Figure 1 is a schematic of a commonly used receiver on a satellite. OBC stands for on board computer. The communication between OBC and satellite receiver can be CAN, rs422, rs485 and other communication methods.

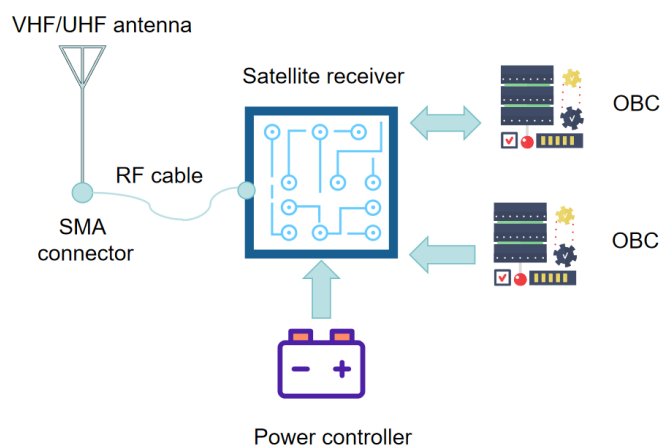

Fig. 1.   A commonly used receiver on a satellite.

Satellite communication modules usually transmit a signal for a very short time, which can reach tens of milliseconds for low Earth Orbit (LEO) satellites on Earth. This is because LEO satellites are relatively close to the ground and the time for signal propagation is short. To improve the data utilization, a signal will include many sensor data packets. The signals will keep coming, and the OBC on the satellite will not only need to parse the packets, but also carry a variety of tasks. Thus, less time is left for anomaly detection. The satellite can't remain unable to parse the data and detect anomalies in real time after receiving it, so the data will keep piling up. Therefore, in our final application scenario, wireless sensor networks and satellite Internet of things complement each other, and their anomaly detection of data requires high real-time performance.

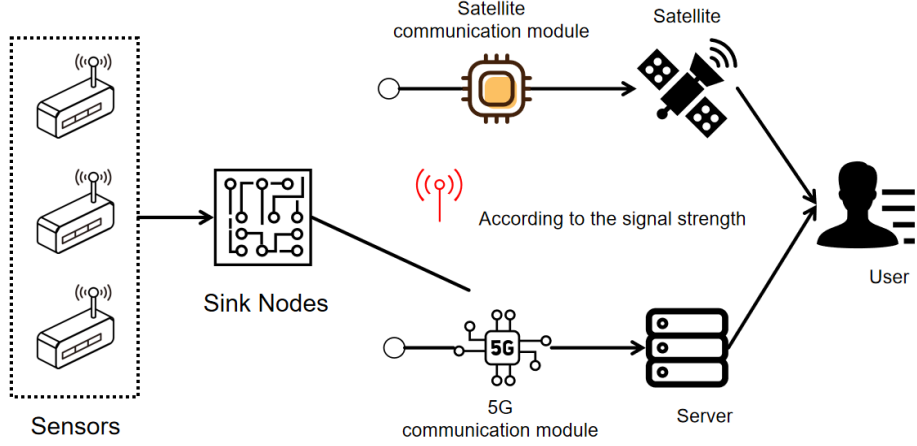

Fig. 2. A smart satellite IoT system.

Satellites and Internet of Things applications are closely intertwined. A common example of such application is a smart satellite IoT system, as illustrated in Figure 2.

## 2.2 Gan-based Anomaly Detection

The basic task of anomaly detection is to identify abnormal points or abnormal sequences within environmental sensor data. While outlier detection is an important aspect, the focus should be on identifying the abnormal sequence itself, which inherently includes abnormal points. Therefore, detecting the abnormal point can lead to the discovery of the abnormal sequence. The size of the abnormal sequence may vary based on parameters such as the time window, step size, and other settings determined during training.

Given a training dataset $X \in R^{M*K}$, where $K$ represents the number of features or dimensions of the variables, and $M$ represents the length of the time series in the training set. The test dataset $Y \in R^{M*K}$, where $N$ represents the length of the time series in the test set. The objective of anomaly detection is to assign multivariate labels to the test dataset (e.g., where 0 indicates an anomaly, 1 indicates normal data for device type 1, 2 indicates normal data for device type 2, 3 indicates normal data for device type 3). Therefore, fundamentally, anomaly detection is a multi-class classification problem based on device identification [10]. While most works treat it as a binary classification problem, here we aim to specifically identify the device category. Thus, we upgrade the problem to a multi-class classification task.

Considering the characteristics of anomaly detection tasks and the significant imbalance between normal and abnormal data, it is common to use normal data (without anomalies) for modeling sequence data. Subsequently, the constructed model is utilized to reconstruct the test data (including abnormal data) for anomaly detection purposes.

To effectively learn X, a multivariate time series is divided into a set of subsequences using a sliding window approach with a window size of w and a step size of s. Here, $x = \{x_i | i = 1, 2, \ldots m\}$, where $x_i \in R^{w*K}$, and $m = \frac{M-w}{s}$ is the number of subsequences. Similarly, $z = \{z_i | i = 1, 2, \ldots m\}$ represents a set of subsequences sampled from the latent space. By feeding x and z into a GAN model, we train the generator and discriminator using a minimax game between the two.

$$\min_{G} \max_{D} V(G, D) = E_{x_i \sim P_{data}}[logD(x_i)] + E_{z_i \sim P_z}[1 - logD(G(z_i))] \qquad (1)$$

Where $P_{data}$ represents the distribution of normal data and $P_z$ represents the distribution of the latent space.

This paper employs the self-attention mechanism to capture correlations among sensor data features. While LSTMs and transformers excel at learning long-term global dependencies, attention mechanisms are more effective at capturing local dependencies. In the field of time series anomaly detection, considering only long-term dependencies is insufficient. Figure 3 demonstrates that the two subsequences (Seq1, Seq2) contain different local information, with Seq2 exhibiting stronger identifiability and containing more critical information. By prioritizing Seq2 and paying closer attention to it, we can capture more useful local features.

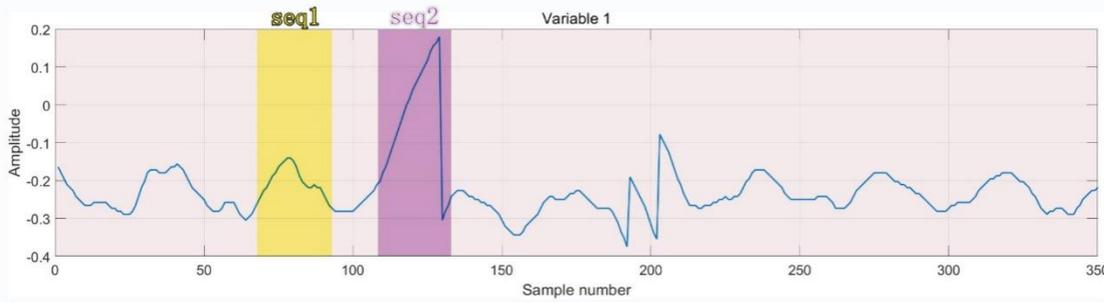

Fig. 3.    The local information contained in a subsequence.

By employing various training techniques, such as learning rate schedulers, and after enough iterations, both the discriminator and generator are effectively trained. GAN-based anomaly detection consists of two stages: the model training phase and the anomaly detection phase. The flow of information during the training process is illustrated in Figure 4.

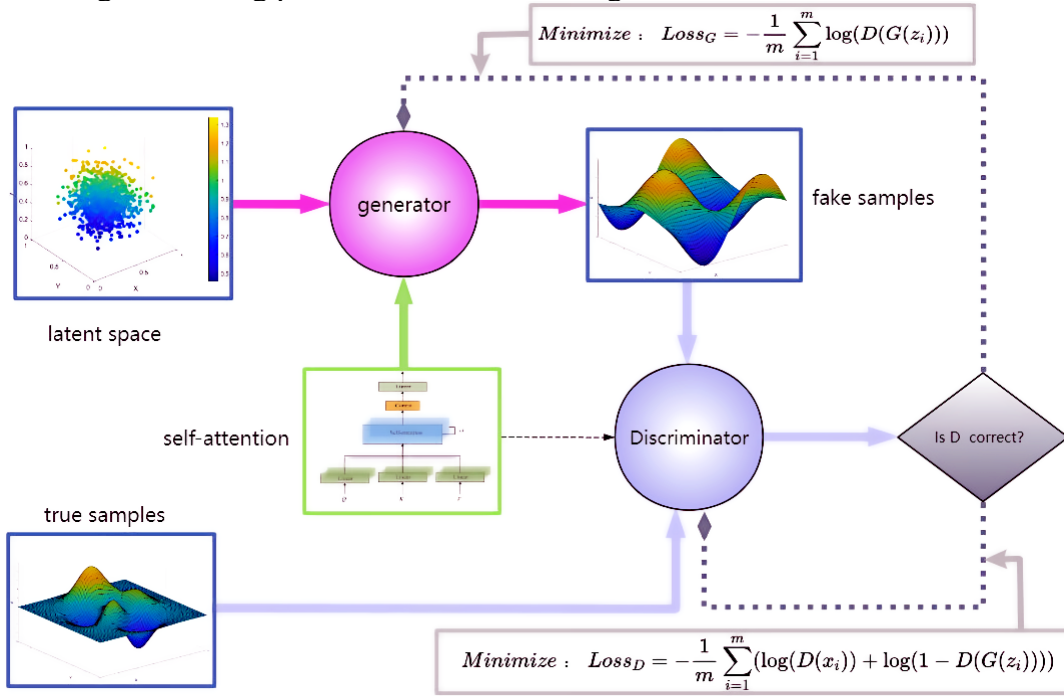

Fig. 4. The flow of information during the training process.

To further uncover potential anomalies, we combine the reconstruction error and the discrimination error to calculate an anomaly score.

**Reconstruction Error:** During the process of reconstructing anomalous information, some information may be lost. Anomalies exhibit a distinct distribution from normal data. Therefore, the error between the test data and the reconstructed data can be utilized to identify anomalies.

**Discrimination Error:** A well-trained discriminator can differentiate between real and anomalous data, making it a direct tool for anomaly detection.

To delve deeper into potential anomalies, we employ the combined reconstruction error to calculate the anomaly score.

$$Res(y_i) = \lambda \|y_i - G(z_i')\| + (1 - \lambda)\|D(y_i) - D(G(z_i'))\| \tag{2}$$

Where $\|*\|$ represents the L2 norm. $z_i'$ represents the optimal one corresponding to $y_i$.

## 3. Experiment

We measured sensor data (sensor readings and execution status) under normal operating conditions for 14 days. Due to the characteristics of enterprise projects, there are many terminal devices, resulting in a very large amount of data itself. Merging multiple edge nodes into a central node of a small area can significantly reduce the cost, but the amount of data that the central node needs to process will be multiplied many times. An average of 4,628,800 pieces of data are aggregated in one day. The device itself is not only responsible for anomaly detection, but also for parsing data, processing data, summarizing data, uploading data to the cloud, and so on, which will also occupy the resources of the central node. Therefore, this device is required to achieve extremely high real-time anomaly detection.

In view of the large amount of data every day and the high similarity between the data, the GAN model will use the statistical point of view to judge the similarity of the data before training, and only take a small part of the data, and a large number of repeated similar data will be discarded (for example, using the mean square error calculation, when the mean square error is small, part of the data can be discarded). To avoid overfitting, we sometimes mean the samples every few steps, which can also speed up training. Also, use some training tricks like the learning rate scheduler. For each day of data, we inserted about 51,440 abnormal data (including cases of accidents and cyberattacks). From these huge data, we analyze, extract... Finally, a total of 716,870 training samples and 179220 testing samples were integrated.

In anomaly detection, we use jetson development board as node device, and TensorRT as inference engine to continuously collect and monitor sensor data. We evaluate the performance of the model using the following metrics, namely precision (pre), Recall (Rec), F1 score.

Table 1. Performance Evaluation

| Method | P(%) | R(%) | F1 |
|---|---|---|---|
| PCA [11] | 25.35 | 22.13 | 23.63 |
| KNN [12] | 13.23 | 11.28 | 12.17 |
| FB [13] | 17.48 | 16.21 | 16.82 |
| CNN | 70.23 | 59.92 | 64.66 |
| LSTM | 67.14 | 40.16 | 50.25 |
| DCGAN | 96.21 | 55.74 | 70.58 |
| MAD-GAN [14] | 98.92 | 65.78 | 79.01 |
| Our model | **99.18** | **70.67** | **82.53** |

As can be seen from Table 1, traditional unsupervised methods: such as PCA, KNN, FB have lost their effect when dealing with high-dimensional data. CNNS and LSTMS also perform poorly due to the lack of labeled data. For a model like MAD-GAN, considering the real-time performance of anomaly detection, we fine-tune the model, resulting in its performance not as good as our model. attention mechanism can be calculated in parallel, so the inference speed is fast, which is suitable for real-time requirements in the embedded field.

## 4. Summary

The paper solves the three challenges mentioned in the abstract using Gans and self-attention with some training techniques. For some small and medium-sized environmental sensor networks, this system is more suitable for anomaly detection tasks of most environmental sensor networks of satellite and Internet of things enterprises and is suitable for real production environments. Running on the smart satellite Internet of things system ,it can save costs, improve computing power, and improve system security.

There are few discussions about real-time in many literatures. Attention mechanism can parallelize the calculation and capture local features. We also use TensorRT inference engine to implement the proposed method, improving real-time performance while maintaining accuracy. We conclude that the combination of GAN and self-attention has great potential and promise in the anomaly detection application field of satellite IoT.